\pdfoutput=1

\documentclass[11pt]{article}

\usepackage[final]{acl}

\usepackage{times}
\usepackage{latexsym}

\usepackage[T1]{fontenc}

\usepackage[utf8]{inputenc}

\usepackage{microtype}

\usepackage{inconsolata}

\usepackage{graphicx}

\usepackage{amsmath,amsfonts,bm}

\def\eqref#1{equation~\ref{#1}}

\def\1{\bm{1}}

\DeclareMathAlphabet{\mathsfit}{\encodingdefault}{\sfdefault}{m}{sl}
\SetMathAlphabet{\mathsfit}{bold}{\encodingdefault}{\sfdefault}{bx}{n}

\usepackage{algorithm}
\usepackage{algorithmic}
\usepackage{setspace}
\usepackage{comment}
\usepackage{booktabs}
\usepackage{hyperref}
\usepackage{icomma}
\usepackage{url}

\newcommand{\Aref}[1]{\hyperref[#1]{Appendix~\ref*{#1}}}

\usepackage[nameinlink]{cleveref}

\usepackage{longtable}
\usepackage{xcolor}
\usepackage{colortbl}
\usepackage{multicol}
\usepackage{multirow}
\usepackage{makecell}
\usepackage{amsmath, amssymb}
\usepackage{mathtools}
\usepackage{enumitem}
\usepackage{subcaption}
\usepackage{float}
\usepackage{graphicx}
\usepackage{caption} 
\usepackage{upgreek}
\usepackage{seqsplit}
\usepackage{color,soul}
\usepackage{arydshln}
\usepackage{xspace}
\usepackage{adjustbox}
\usepackage{enumitem}
\usepackage{booktabs} 
\usepackage[most]{tcolorbox}

\definecolor{Red}{rgb}{0.6,0,0}
\definecolor{Blue}{rgb}{0,0,0.8}
\definecolor{Green}{rgb}{0,0.4,0.7}
\definecolor{mountainmeadow}{rgb}{0.19, 0.73, 0.56}
\definecolor{crimson}{rgb}{0.86, 0.08, 0.24}
\definecolor{darkblue}{rgb}{0.0, 0.0, 0.55}

\hypersetup{
    colorlinks = true,
    citecolor = mountainmeadow,
    linkcolor = crimson,
    linktoc = all,
    urlcolor = darkblue,
}

\newcommand{\highlight}[1]{{\color{crimson}{#1}}}

\newtcolorbox{Box1}[2][]{
    lower separated=false,
    colback=white!80!gray,
    colframe=white, fonttitle=\bfseries,
    colbacktitle=white!50!gray,
    coltitle=black,
    enhanced,
    attach boxed title to top left={xshift=0.5cm,yshift=-2mm},
    title=#2,#1
}

\title{Efficient Real-time Refinement of Language Model Text Generation}

\author{
    Joonho Ko \;\;
    Jinheon Baek \;\; 
    Sung Ju Hwang \\
    KAIST\\
    \texttt{\{joonho.ko, jinheon.baek, sungju.hwang\}@kaist.ac.kr}
}

\begin{document}
\maketitle
\begin{abstract}

Large language models (LLMs) have shown remarkable performance across a wide range of natural language tasks. However, a critical challenge remains in that they sometimes generate factually incorrect answers. To address this, while many previous work has focused on identifying errors in their generation and further refining them, they are slow in deployment since they are designed to verify the response from LLMs only after their entire generation (from the first to last tokens) is done. Further, we observe that once LLMs generate incorrect tokens early on, there is a higher likelihood that subsequent tokens will also be factually incorrect. To this end, in this work, we propose \textbf{Streaming-VR} (\textbf{Streaming} \textbf{V}erification and \textbf{R}efinement), a novel approach designed to enhance the efficiency of verification and refinement of LLM outputs. Specifically, the proposed Streaming-VR enables on-the-fly verification and correction of tokens as they are being generated, similar to a streaming process, ensuring that each subset of tokens is checked and refined in real-time by another LLM as the LLM constructs its response. Through comprehensive evaluations on multiple datasets, we demonstrate that our approach not only enhances the factual accuracy of LLMs, but also offers a more efficient solution compared to prior refinement methods. 

\end{abstract}

\section{Introduction}

Large language models (LLMs)~\citep{gpt4, mistral, llama3} have demonstrated significant advancements across various tasks, such as question answering (QA)~\citep{hotpotqa, nq, eli5, ambigqa} and their more complex real-world applications, which are even supported by information retrieval (IR) for accurately generating answers~\citep{ance, dpr, gtr}. However, LLMs still face notable limitations like hallucinations, mainly due to the incorrect or outdated knowledge of the model itself~\citep{troubling} and the wrong application and generalization of memorized or retrieved knowledge~\citep{hallucination, inevitable}.

Previous approaches have sought to mitigate these inaccuracies by augmenting LLMs with external knowledge sources~\citep{realm, rag, sail}. However, these methods often face challenges in maintaining faithfulness, as they may retrieve information that is either ungrounded or irrelevant to the context. To this end, in the realm of error identification and verification, recent research has highlighted the challenges LLMs face in accurately detecting and correcting mistakes~\citep{check}.

\begin{figure*}[t]
\centering
\includegraphics[width=0.975\textwidth]{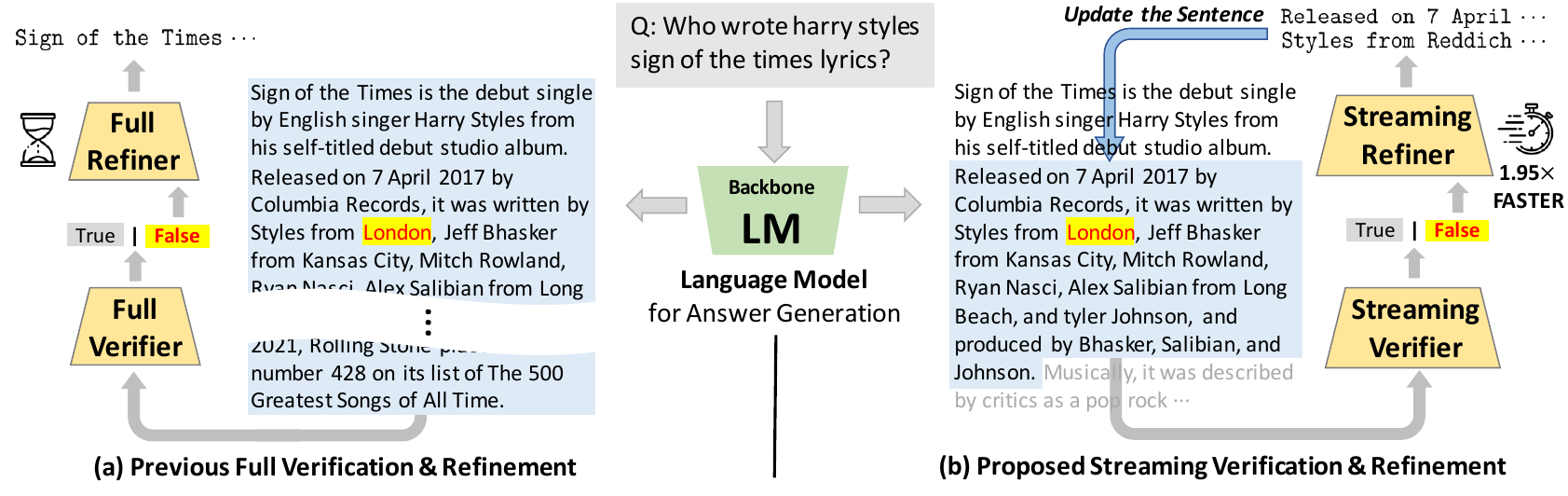}
\caption{\small \textbf{(a)}: Previous verify-and-refine framework after the entire answer generation. \textbf{(b)}: Our proposed method, \textbf{Streaming-VR}, that verifies intermediate answers in sentence-level and refine them if identified as error, with remarkable efficiency.}
\label{fig:concept}
\end{figure*}

However, traditional methods~\citep{text-editing, lm-critic, self-refine} have a couple of challenges. First, they are inefficient. They focus mainly on identifying and correcting misinformation only after the complete answer has been generated. This approach not only delays error detection but also requires re-evaluating the entire text, which is computationally expensive and time-consuming. Second, cascading errors. LLMs generate text sequentially, predicting one token at a time based on the preceding context. An early error in this sequence can propagate through subsequent tokens, compounding inaccuracies throughout the response. This error propagation makes it even more challenging to correct misinformation effectively, especially when early mistakes lead to increasingly complex or numerous errors to the overall response. These challenges highlight the critical need for intermediate corrections to prevent further inaccuracies throughout the response.

In this work, we propose \textbf{Streaming} \textbf{V}erification and \textbf{R}efinement, in short \textbf{Streaming-VR}, which is a method designed to address the issue of error propagation in LLM-generated text. As visually depicted in \Cref{fig:concept}~\highlight{(b)}, Streaming-VR evaluates model-generated answers in real-time, identifying the entire token sequence and correcting only if its subset is wrong. By employing an external verification model, Streaming-VR verifies errors during the generation process, detects inaccuracies in the newly generated sequence of tokens, and promptly corrects them. Because rectification occurs immediately after verification and runs concurrently with text generation, Streaming-VR significantly enhances efficiency and improves the factual accuracy of model outputs. Our experimental results show that when LLMs generate incorrect tokens early in a sequence, it substantially increases the likelihood of subsequent sentences being factually inaccurate. Specifically, approximately 37.6\% of the answers in various settings were found to contain factual inaccuracies caused by error propagation (early erroneous tokens), highlighting the critical importance of employing Streaming-VR.

We validate the effectiveness and efficiency of Streaming-VR experimentally on two benchmark datasets: achieving approximately 39.8\% and 31.5\% higher efficiency for ASQA~\citep{asqa} and QuoteSum~\citep{quotesum} in average, respectively. Also, Streaming-VR is approximately 1.95$\times$ faster than Full-VR with comparable answer quality. We have employed Mistral 7B~\citep{mistral}, LLaMA-3.1 8B~\citep{llama3}, and GPT-4o~\citep{gpt4}.
\section{Related Work}

\paragraph{Large Language Models}
Recent advancements in language models (LMs) \citep{gpt, bert, roberta, t5} and LLMs with billions of parameters have led to significant improvements in performance across various natural language tasks. Since LMs cannot memorize or learn every real-world knowledge, several studies have explored methods to enhance their capabilities by leveraging external knowledge sources like retrieval-augmented generation~\citep{rag}, for knowledge-intensive tasks. Despite the assistance of external knowledge, models often generate incorrect answers due to the failure of factual recall~\citep{hallucination} as they may not succeed in retrieving or applying the relevant knowledge appropriately, and generalizing memorized knowledge accurately.

To address this issue, recent research has focused on verifying the relevance and accuracy of retrieved knowledge using separate verification mechanisms~\citep{kalmv}. Additionally, methods for generating answers through on-demand retrieval of external information, employing special retrieval tokens, followed by critiquing the outputs to improve their quality, have been explored~\citep{self-rag}. A dynamic retrieval process that determines both when and what to retrieve during answer generation~\citep{flare} has demonstrated notable improvements in knowledge-intensive tasks. This is particularly significant as the retrieve-and-generate paradigm faces significant challenges in generating lengthy texts, primarily due to difficulties in maintaining coherence and consistency. Retrieved knowledge is often fragmented and lacks contextual integration, while static retrieval methods fail to adapt dynamically to evolving text, leading to disjointed or repetitive outputs. Future research could address these issues through iterative retrieval mechanisms that refine knowledge during generation, advanced reasoning capabilities to synthesize information from multiple sources, and hierarchical retrieval strategies~\citep{adaptive-rag} that organize information at different levels of granularity and difficulty leveraging an external query complexity classifier.

\paragraph{Language Model Verification and Refinement}
Other than the misinformation induced by wrong knowledge, LLM itself often generates plausible but incorrect texts~\citep{snowball} (\textit{i.e.}, hallucination). Thus, evaluating the factuality ~\citep{fever, factscore} of LLM outputs correcting inaccuracies has emerged as an important topic. Various approaches explore methods to enhance the factual accuracy of model responses and develop robust fact-checking or answer-verifying models. For instance, \citet{cove} generates a series of independent questions to check the factual claims made in the model response, followed by synthesizing the answers from the verification step. Beyond evaluating or verifying the faithfulness of LLM answers, answer-correction has also become a prominent area of focus in various fields. Iterative refinement is well known to be helpful for improving generative contents of natural language~\citep{self-refine} and code~\citep{text-editing, lm-critic} autonomously, but is limited to the final outcome after waiting for the whole generation to be done.

On the other hand, \citet{verify-step} demonstrates the effectiveness of process supervision by focusing on each step of the reasoning process, and allowing the model to identify and correct errors in the middle. It emphasizes the importance of intermediate verification in complex multi-step reasoning tasks like mathematical problem solving, where a single error can derail the entire answer. Also, \citet{self-correct} employs an online training procedure for a separate corrector to learn from feedback on intermediate outputs. Nevertheless, LMs are capable of correcting errors only when their locations are identified~\citep{llm-mistake} exactly, which poses a bottleneck in improving self-correction capabilities. Furthermore, \citet{self-correct-reasoning} have demonstrated through experimental analyses that current LLMs struggle to self-correct their reasoning without external feedback, often resulting in degraded performance after attempting self-correction. Alternatively, \citet{training-verifiers, shepherd} utilize a trained critique model or verifier to correct errors on responses through their feedback. In addition, \citet{critic} show that verification and correction can be done effectively by interacting with diverse external tools. In contrast to the previous works, which have to wait for the entire answer generation or are limited to the inherent answering ability, we propose a novel method with an external model that refines the specific intermediate sentence of an answer identified as incorrect, with higher efficiency.

\section{Method}
\label{method}

\subsection{Preliminaries}
We begin with preliminaries, explaining Large Language Models and the traditional verify-and-refine approach, Full Verification and Refinement.

\paragraph{Large Language Models}
Let us define the process of generating an answer $\boldsymbol{a}$ to a given question $\boldsymbol{q}$ as a function: $\boldsymbol{a} = \texttt{LLM}(\boldsymbol{q})$.

For the real-time sentence-level verification and refinement, we also analyze the individual sentences in the answer. To elaborate, an answer $\boldsymbol{a}$ is structured as a sequence of $n$ sentences, expressed as $\boldsymbol{a} = \left[ s_1, s_2, \cdots, s_n \right]$, where the notation $\left[ \cdot \right]$ signifies concatenation in the specified order. To facilitate real-time correction of incorrect sentences within intermediate answers, we define the intermediate answer at a certain step $t$ ($t \geq 1$) as $\boldsymbol{a}_{\leq t} = \left[ s_1, \cdots, s_t \right]$ containing $t$ sentences in total. Note that this can also be expressed as $\boldsymbol{a}_{\leq t} = \left[ \boldsymbol{a}_{\leq t-1}, s_t \right]$, where $s_t$ is the most recently generated sentence in a streaming setup. We initialize $\boldsymbol{a}_{\leq 0}$ as an empty string for coherence.

In QA systems that incorporate external knowledge, such as in retrieval-augmented generation (RAG), or examples as in in-context learning (ICL), the answering process differs slightly. Formally, let $\boldsymbol{d}$ denote the external knowledge or example retrieved from the source $\mathcal{D}$. The retrieval is performed using a dedicated retrieval model \texttt{Retriever}, for a given query $\boldsymbol{q}$, defined as: $\boldsymbol{d} = \texttt{Retriever}\left( \boldsymbol{q}; \mathcal{D} \right)$. This process involves ranking the retrieved data based on its relevance or similarity to the given query. After the related documents are retrieved for RAG or ICL, we now incorporate them as input to the LLMs as: $\boldsymbol{a} = \texttt{LLM}(\boldsymbol{q}, \boldsymbol{d})$.

\paragraph{Full-VR}
The simplest traditional approach for verifying and refining LLM answers, namely Full-VR (Full Verification and Refinement), is the most common strategy for improving them just by re-generating the entire responses if identified as incorrect. While many previous works~\citep{lm-critic, training-verifiers, self-correct} achieve significant improvements through supplementary techniques, we focus solely on the vanilla setting, for a direct efficiency comparison without any additional methods designed to increase the factual accuracy of answers. And finally, the overall Full-VR pipeline is expressed as follows for a given query $\boldsymbol{q}$ and its answer $\boldsymbol{a} = \texttt{LLM}\left(\boldsymbol{q}\right)$:
\begin{align}
    \tilde{\boldsymbol{a}} = 
        \begin{cases}
            \boldsymbol{a} & \textit{if} \quad o=\texttt{True}\\ \notag
            \texttt{Refiner}\left(\boldsymbol{a}\right) & \textit{if} \quad o=\texttt{False}
        \end{cases} 
\end{align}
where $o = \texttt{Verifier}\left(\boldsymbol{a}\right)$ is the verification output, and $\tilde{\boldsymbol{a}}$ is the final output of Full-VR.

\subsection{Streaming Verification and Refinement}
Our approach is structured in the following steps during the generation of answers: 1) Streaming-Verification, and 2) Streaming-Refinement if necessary (for the sentence identified as an error) and then go back to 1). We formulate the overall framework of Streaming-VR for a given query $\boldsymbol{q}$ and the $t$-th sentence $s_t \in \texttt{LLM}\left(\boldsymbol{q}\right)$ in its answer, as follows:
 \begin{equation}
    \tilde{s}_{t} = 
        \begin{cases}
            s_t & \textit{if} \quad o_t=\texttt{True}\\ \notag
            \texttt{Refiner}\left(s_t\right) & \textit{if} \quad o_t=\texttt{False}
        \end{cases}
\end{equation}
where $o_t = \texttt{Verifier}\left(\left[\tilde{\boldsymbol{a}}_{\leq t-1}, s_t \right]\right)$ is the verification output, and $\tilde{s}_{t}$ is the new sentence output of Streaming-VR at a certain step $t$. Note that the refinement model, \texttt{Refiner} takes into account the whole context of previously verified and (may have been) refined sentences, $\tilde{\boldsymbol{a}}_{\leq t-1}=\left[\tilde{s}_1, \cdots, \tilde{s}_{t-1}\right]$. After processing all the sentences by Streaming-VR, the final refined answer output should be in the form as follows: $\tilde{\boldsymbol{a}}=\left[\tilde{s}_1, \cdots, \tilde{s}_n \right]$.

The answer verification relies on the verifier's output, $o_t = \texttt{Verifier}\left(\boldsymbol{a}_{\leq t}\right)$ such that $o_t \in \left\{\texttt{True}, \texttt{False}\right\}$. We utilize a fine-tuned LLM to determine whether the input is \texttt{True} or \texttt{False} by evaluating the factuality of the generated answers at the sentence-level. To this end, we augment training data with true- and false-labeled sentences, as there is no proper question answering dataset labeled accurately with unit-level (\textit{e.g.} sentence-level) answers for our streaming-verifier. The augmented sentences are made from the provided reference answer data by rephrasing it for \texttt{True} and adding wrong information for \texttt{False} by GPT-4o~\citep{gpt4} with the specific prompt as in Implementation Details (\Aref{app:augmentation_prompt}). To suit real-time verification scenarios, we split the answer data into individual sentences using NLTK~\citep{nltk}. These sentences are concatenated incrementally in their original order to form intermediate answers $\left\{\boldsymbol{a}_{\leq 1}, \cdots, \boldsymbol{a}_{\leq t} \right\}$, ensuring that \texttt{False}-labeled sentences only appear at the end, never in the middle. This design allows the streaming verifier to focus on determining the factuality of the newly-generated sentence at the end. To further enhance the training process, a special sentence-separation token, \texttt{[SEP]}, is inserted right before the last sentence in each intermediate answer, formatted as $\left[s_1, s_2, \cdots, \texttt{[SEP]}, s_t \right]$ for a certain stage $t$. This setup allows a model to be trained to verify the last sentence along with the context from the preceding \texttt{True}-labeled paragraph in the train set.

To facilitate a real-time scenario with conventional language models, we provide the entire prompt given for answering the test query to the refinement model. Additionally, we only include the retrieved passages or few-shot examples given to the generation prompt, without incorporating any extra information from external knowledge sources for refinement. This strategy ensures that the contextual information relevant to the intermediate generation processes is fully incorporated. Furthermore, as the intermediate answers are refined, they must be updated to reflect the newly refined preceding sentences, thereby enabling a continuous and coherent streaming refinement process.

\section{Experiments}

\subsection{Datasets and Evaluation Metrics}\label{datasets and metrics}
We use two different datasets to evaluate the effect of Streaming-VR, especially for multi-answer questions, which require well-grounded responses to assess the trustworthiness of QA systems.

\begin{table*}[t]
\centering
\setlength{\tabcolsep}{3.5pt}
\scriptsize
\caption{\small Results of Streaming-Verification by \textbf{Mistral 7B} and Refinement by \textbf{GPT-4o} on \textbf{ASQA} and \textbf{QuoteSum} for three different backbone response models. $\mathcal{T}_\text{Ref}$ indicates the number of newly-generated tokens for refinement.}
    \centering
        \begin{tabular}{l ccccc ccccc}
            \toprule
            \multicolumn{11}{c}{\small \textbf{ASQA}} \\
            \midrule[0.5pt]
            & \multicolumn{5}{c}{\textbf{Closed-Book}} & \multicolumn{5}{c}{\textbf{Open-Book with 5 Passages}} \\
            \cmidrule(lr){2-6} \cmidrule(lr){7-11}
            Method & ROUGE-L & Disambig-F1 & DR & $\mathcal{T}_\text{Ref}$ & Efficiency $\uparrow$ & ROUGE-L & Disambig-F1 & DR & $\mathcal{T}_\text{Ref}$ & Efficiency $\uparrow$\\
            
            \midrule[0.5pt]
            Mistral 7B & $33.6$ & $20.7$ & $26.4$ & $-$ & $-$
            & $36.4$ & $31.2$ & $33.7$ & $-$ & $-$ \\
            + Full-VR & $35.3$ & $29.6$ & $32.3$ & $113.6$ & \multirow{2}{*}{$39.6\%$}
            & $36.6$ & $33.9$ & $35.2$ & $101.8$ & \multirow{2}{*}{$26.9\%$}   \\
            + \textbf{Streaming-VR} & $35.2$ & $29.6$ & $32.3$ & $68.6$ &  
            & $36.9$ & $33.7$ & $35.3$ & $74.4$ &  \\
            
            \midrule[0.5pt]
            LLaMA-3.1 8B & $34.0$ & $23.7$ & $28.4$ & $-$ & $-$ 
            & $36.6$ & $31.7$ & $34.1$ & $-$ & $-$ \\
            + Full-VR & $35.2$ & $29.4$ & $32.2$ & $117.4$ & \multirow{2}{*}{$45.8\%$}
            & $37.0$ & $34.2$ & $35.6$ & $106.8$ & \multirow{2}{*}{$42.1\%$} \\
            + \textbf{Streaming-VR} & $35.3$ & $29.4$ & $32.2$ & $63.6$ &
            & $36.8$ & $34.0$ & $35.4$ & $61.9$ &  \\
        
            \midrule[0.5pt]
            GPT-4o & $36.6$ & $34.8$ & $35.7$ & $-$ & $-$ 
            & $37.1$ & $35.0$ & $36.0$ & $-$ & $-$ \\
            + Full-VR & $35.2$ & $29.6$ & $32.3$ & $100.4$ & \multirow{2}{*}{$38.3\%$}
            & $37.0$ & $33.9$ & $35.4$ & $116.1$ & \multirow{2}{*}{$46.0\%$} \\
            + \textbf{Streaming-VR} & $35.3$ & $29.4$ & $32.2$ & $61.9$ &  
            & $36.9$ & $33.9$ & $35.4$ & $62.7$ &  \\

            \toprule
            \multicolumn{11}{c}{\small \textbf{QuoteSum}} \\
            \midrule[0.5pt]
            & \multicolumn{5}{c}{\textbf{Zero-Shot}} & \multicolumn{5}{c}{\textbf{Five-Shots}} \\
            \cmidrule(lr){2-6} \cmidrule(lr){7-11}
            Method & ROUGE-L & Sem-F1 & SEMQA & $\mathcal{T}_\text{Ref}$ & Efficiency $\uparrow$ & ROUGE-L & Sem-F1 & SEMQA & $\mathcal{T}_\text{Ref}$ & Efficiency $\uparrow$ \\
        
            \midrule[0.5pt]
            Mistral 7B & $37.5$ & $39.0$ & $38.2$ & $-$ & $-$
            & $46.8$ & $51.8$ & $50.1$ & $-$ & $-$ \\
            + Full-VR & $38.1$ & $39.0$ & $38.5$ & $101.3$ & \multirow{2}{*}{$25.8\%$}
            & $57.6$ & $49.0$ & $53.1$ & $72.5$ & \multirow{2}{*}{$24.3\%$} \\
            + \textbf{Streaming-VR} & $37.9$ & $39.0$ & $38.4$ & $75.2$ &
            & $57.5$ & $48.9$ & $52.9$ & $54.9$ \\
            
            \midrule[0.5pt]
            LLaMA-3.1 8B & $43.3$ & $38.9$ & $41.0$ & $-$ & $-$
            & $59.1$ & $61.2$ & $60.1$ & $-$ & $-$ \\
            + Full-VR & $47.6$ & $39.0$ & $43.1$ & $154.3$ & \multirow{2}{*}{$31.2\%$}
            & $60.7$ & $62.1$ & $61.4$ & $84.1$ & \multirow{2}{*}{$30.0\%$}\\
            + \textbf{Streaming-VR} & $47.5$ & $39.0$ & $43.0$ & $106.1$ & 
            & $60.7$ & $62.3$ & $61.5$ & $58.9$ & \\
        
            \midrule[0.5pt]
            GPT-4o & $60.3$ & $39.0$ & $48.5$ & $-$ & $-$
            & $65.8$ & $54.7$ & $60.0$ & $-$ & $-$ \\
            + Full-VR & $60.2$ & $39.0$ & $48.5$ & $60.7$ & \multirow{2}{*}{$26.7\%$}  
            & $65.8$ & $54.7$ & $60.0$ & $78.9$ & \multirow{2}{*}{$42.2\%$} \\
            + \textbf{Streaming-VR} & $60.0$ & $39.0$ & $48.4$ & $44.5$ &
            & $65.3$ & $54.7$ & $59.8$ & $45.6$ \\
            \bottomrule
        \end{tabular}
    \label{tab:main_result}
\end{table*}

\textbf{ASQA}~\citep{asqa} is a challenging dataset serving as a bridge between factoid and long-form QA tasks by addressing ambiguous questions that can have multiple correct answers depending on their interpretation. It is composed of 4,353 and 948 questions in the train and dev sets, respectively, while the test set is not publicly available. So we use the dev set as our test set here. ASQA provides the reference long-form answers for every question, which originate from AmbigQA~\citep{ambigqa}, the ambiguous questions subset of questions from NQ~\citep{nq}. In this paper, to evaluate the quantitative performance of methods on ASQA, we follow the official metrics and report: Disambiguous-Rouge (DR) as the overall score, which combines ROUGE-L (R-L)~\citep{rouge} for text quality and Disambig-F1 (Dis-F1; QA accuracy score based on RoBERTa large~\citep{roberta}) for factual correctness.

To evaluate the consistent impact of Streaming-VR also in a retrieval-augmented generation (RAG) setting, as in the original ASQA paper~\citep{asqa}, we perform experiments using retrieved documents. Specifically, we use the top-$k$ documents ranked by semantic similarity between the query and external documents for open-book answer generation on the ASQA dataset. These documents, retrieved from the Wikipedia corpus (2018-12-20 snapshot) using GTR-XXL~\citep{gtr}, are provided by the LLM citation benchmark ALCE~\citep{alce}.

\textbf{QuoteSum}~\citep{quotesum} is also a difficult question answering dataset for Semi-Extractive Multi-source Question Answering (SEMQA), a task designed to assess the comprehensive answering ability by summarizing information from multiple sources. Specifically, SEMQA requires models to generate a response that integrates verbatim factual spans extracted from input sources along with supplementary non-factual text connecting them, thereby ensuring a cohesive answer. QuoteSum is made up of 4,009 semi-extractive answers to 1,376 unique questions from PAQ~\citep{paq} and NQ. For the quantitative evaluation on QuoteSum, we follow the official metrics and report: ROUGE-L, Sem-F1 for answer extraction quality, and overall SEMQA score, where they do not require any model-based evaluations.

Building on the original evaluation of QuoteSum~\citep{quotesum}, we further conduct a quantitative assessment of the variants of few-shot models. Specifically, we use a dynamic prompt with top-$k$ examples for each question in the test set, as provided in the original paper. These examples are retrieved from the training set by selecting the passages whose queries are most similar to the target test query, based on the cosine similarity between their sentence embeddings~\citep{sentence-t5}.

\begin{figure*}[ht]
    \normalsize
    \centering
    \resizebox{0.9\linewidth}{!}{
    \begin{tabular}{ccc}
        \includegraphics[width=0.6\linewidth]{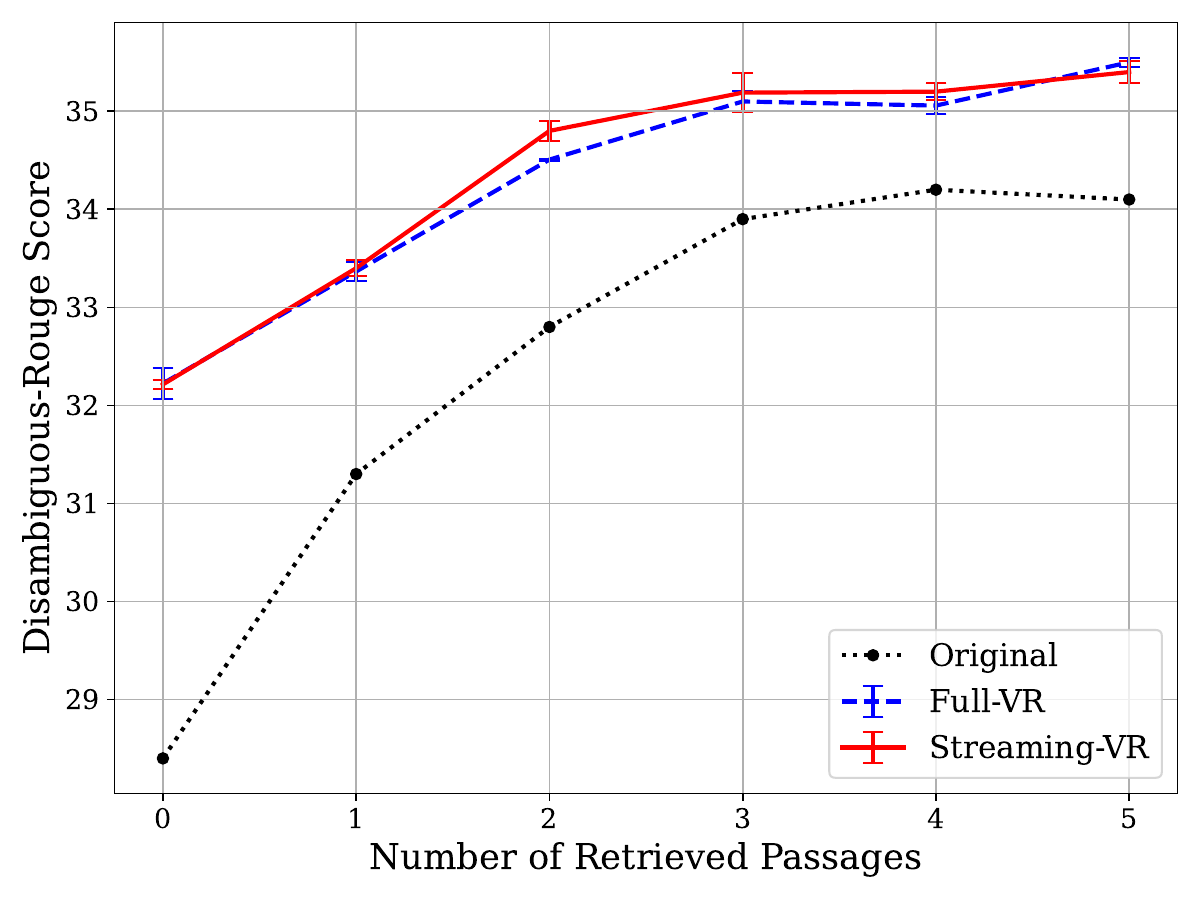} &
        \includegraphics[width=0.6\linewidth]{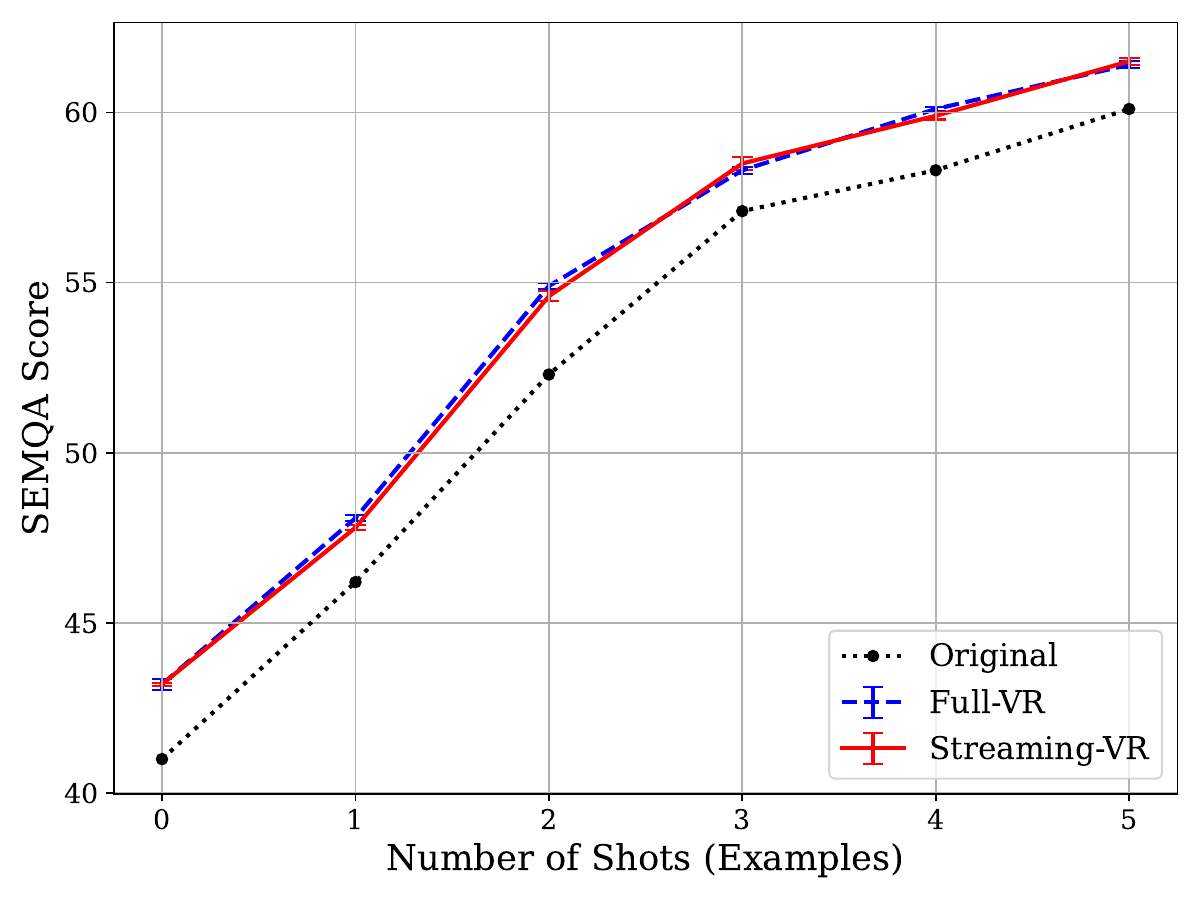} \\
        \textbf{\phantom{,,,,,,,}(a) Performance on ASQA} &
        {\textbf{\phantom{,,,,,,,,}(b) Performance on QuoteSum}} \\
    \end{tabular}
    }
\caption{\small Performance comparison on various RAG and ICL settings.}
\label{fig:performance}
\end{figure*}

\subsection{Analyses on Efficiency}
In addition to evaluating the quality and factual accuracy of model responses, we also measure token count to assess the efficiency of each method. Since our experiments rely on models accessed through the HuggingFace~\citep{huggingface} API, it was not feasible to implement simultaneous execution of the verifier and refiner alongside the answering model, as would occur in real-world applications. Consequently, we analyze the inference cost (\textit{i.e.}, the number of tokens) per model for each method. This metric is crucial as the number of refined tokens directly affects the LLM user's waiting time for response corrections. To quantify the efficiency, we define the efficiency of Streaming-VR relative to Full-VR, taking a cue from the thermal efficiency in thermodynamics, which is formulated as: $\text{(Efficiency)}:= \tfrac{\text{benefit}}{\text{cost}} = 1 - \tfrac{\mathcal{T}_{\text{S}}}{\mathcal{T}_{\text{F}}}$. Here, $\mathcal{T}_{\text{S}}$ and $\mathcal{T}_{\text{F}}$ represent the average number of generated tokens in the refinement phase per answer for Streaming-VR and Full-VR, respectively.

It should be emphasized that the tokens being verified are identical for both methods. Consider an answer with $N$ sentences, where each sentence contains $\mathcal{T}_i$ tokens $\left(i = 1, \dots, N\right)$. Full-VR processes all $\sum_{i=1}^{N} \mathcal{T}_i$ tokens in a single step, whereas Streaming-VR verifies sentence segments sequentially, processing $\mathcal{T}_i$ tokens at step $i$ from $i=1$ to $N$. Despite this difference in approach, both methods process the same total number of tokens, resulting in identical overall verification costs, irrespective of the number of verifier invocations.

\begin{figure*}[t!]
    \normalsize
    \centering
    \resizebox{0.9\linewidth}{!}{
    \begin{tabular}{ccc}
        \includegraphics[width=0.6\linewidth]{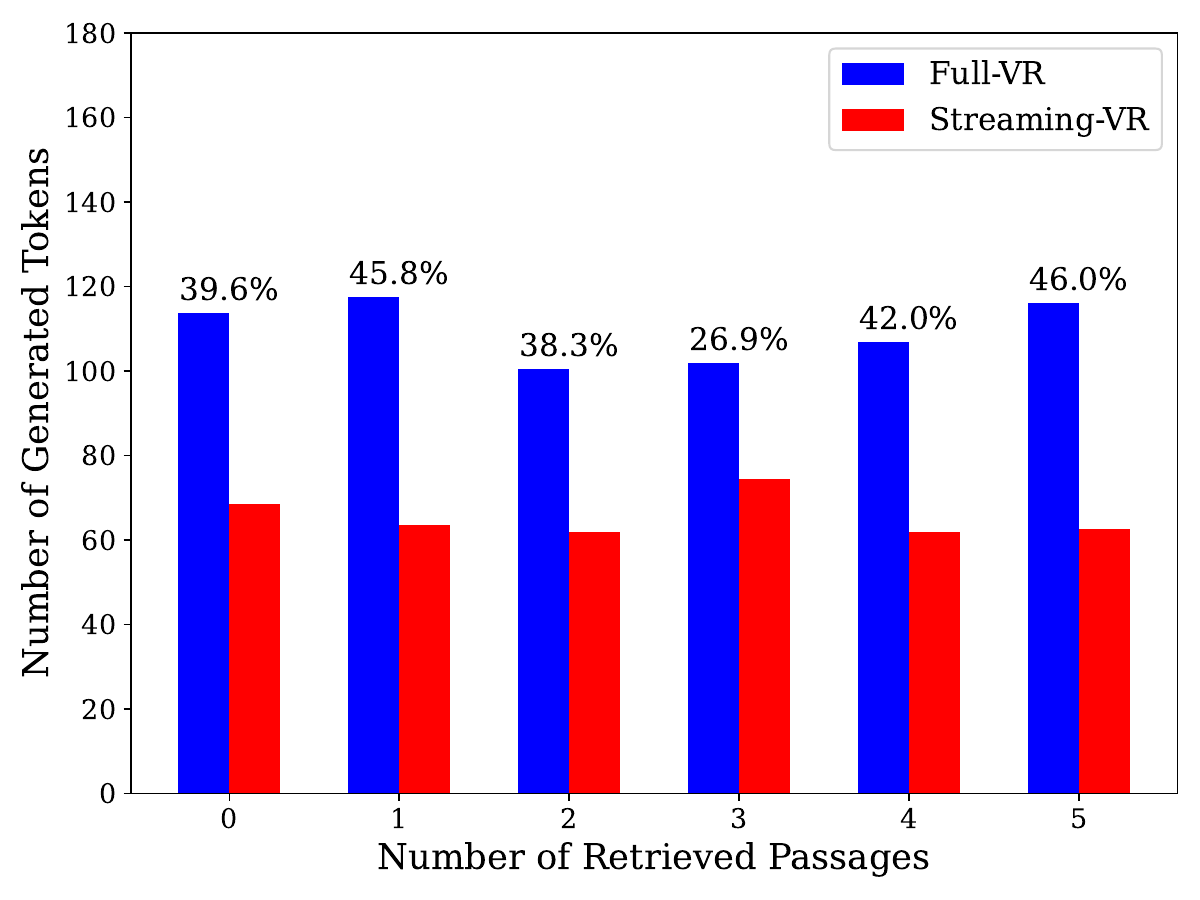} &
        \includegraphics[width=0.6\linewidth]{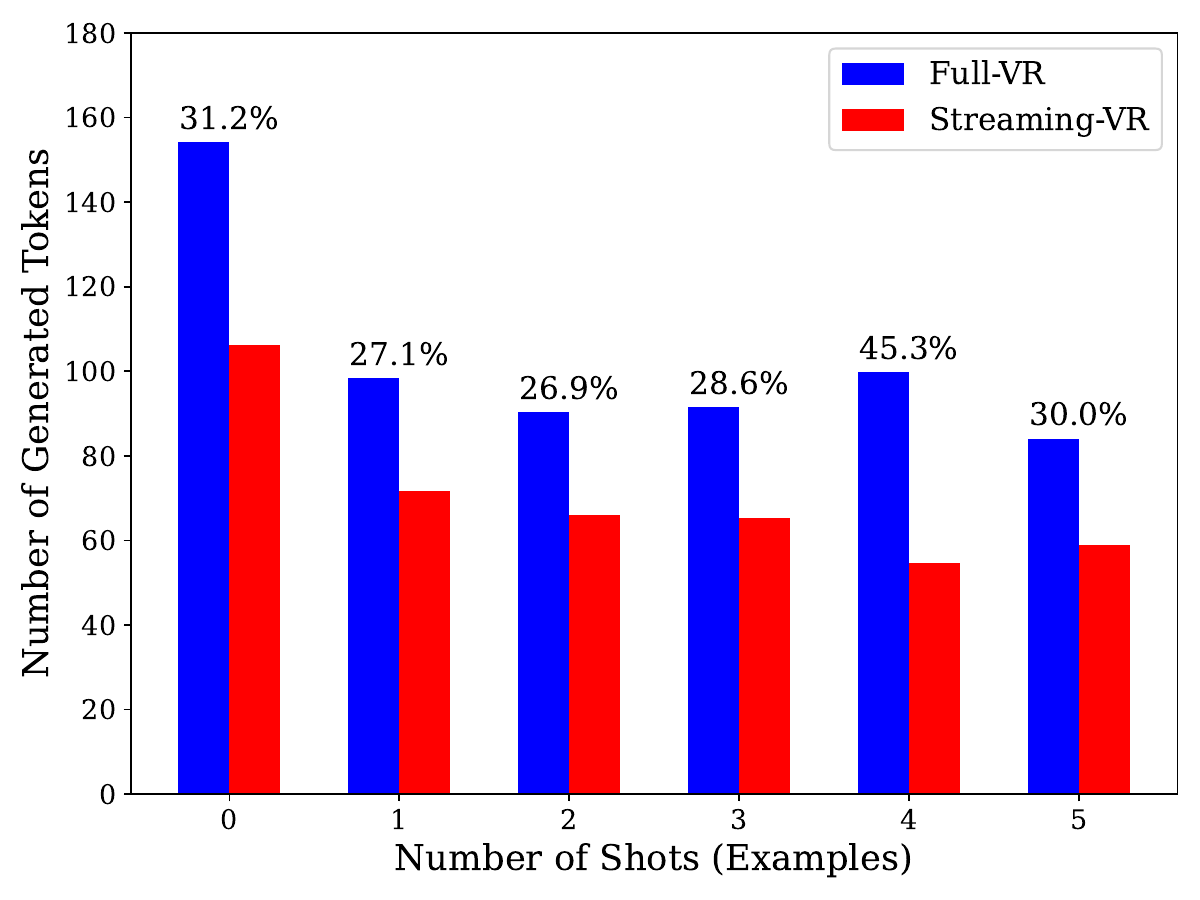} \\
        \textbf{\phantom{,,,,,,,}(a) Efficiency on ASQA} &
        {\textbf{\phantom{,,,,,,,,}(b) Efficiency on QuoteSum}} \\
    \end{tabular}
    }
\caption{\small Efficiency comparison on various RAG and ICL settings. The numbers on top of the bars are their efficiency values.}
\label{fig:efficiency}
\end{figure*}

\subsection{Experimental Results and Analyses}
\label{results}

\paragraph{Streaming-VR delivers higher efficiency while maintaining its performance.}
We conduct a series of experiments on the ASQA and QuoteSum datasets to quantitatively evaluate the efficiency and effectiveness of two approaches: Streaming-VR and Full-VR. For this comparison, we first segment the model-generated answers for each test query into individual sentences, treating these sequentially arranged sentences as distinct intermediate answers. Using Streaming-VR, we verify and refine each intermediate answer in real-time, enabling dynamic adjustments as responses are generated. In contrast, Full-VR serves as the baseline, where the entire answer is verified and refined only after the complete sequence has been generated, processing the output in a single pass from start to finish. Note that for Full-VR, we utilize shared verification results of Streaming-VR: an answer is deemed incorrect if it contains at least one erroneous token in the overall context. By comparing Streaming-VR and Full-VR, we aim to demonstrate the advantages of real-time refinement in improving both answer quality and efficiency.

The main results on ASQA and QuoteSum are summarized in \Cref{tab:main_result}. Both Full-VR and Streaming-VR employ Mistral 7B as the verifier and GPT-4o as the refiner across three different backbone models (Mistral 7B, LLaMA-3.1 8B, and GPT-4o) for answer generation, as indicated in the method column. Across all response models, the final outcomes after verification-and-refinement converge to similar scores, indicating that the overall quality and faithfulness of the answers are largely determined by the refinement model.

However, we observe a notable performance decline when GPT-4o is used as the backbone for answer generation. Both Full-VR and Streaming-VR with GPT-4o lead to significant drops in Disambig-F1 on ASQA, a key metric for assessing the informativeness of long-form answers, and no other improvements on scores of QuoteSum. These results suggest that GPT-4o, which already generated high-quality answers, may be susceptible to over-correction during the refinement process, reducing the overall effectiveness of the responses. This finding highlights a broader trend: refining answers with the same model used for generation—even a powerful model like GPT-4o—may not improve performance and can even degrade it. For applications like large-scale data analysis or high-frequency user requests handling thousands or millions of queries daily, or individual users requiring detailed, lengthy responses, relying on expensive models like GPT-4o for both generation and refinement can quickly exceed budgetary constraints. Therefore, Streaming-VR, which uses a more cost-effective model for response generation and GPT-4o solely for refinement, emerges as a more practical and economical solution.

To assess the consistent efficacy of Streaming-VR across various settings of RAG and ICL for answer generation, we conduct additional experiments as visualized for performance in~\Cref{fig:performance} and for efficiency in~\Cref{fig:efficiency}. The answers are generated by LLaMA-3.1 8B, verified by Mistral 7B and refined by GPT-4o on both datasets. The results show Streaming-VR's competitive performance compared to Full-VR. Streaming-VR consistently outperforms the initial answers without refinement and achieves comparable results to Full-VR. It also illustrates that Streaming-VR delivers results on par with Full-VR across all retrieved passages and example shot counts, offering performance improvements over the unrefined original response outputs of the language model.

\begin{table}[t]
\centering
\setlength{\tabcolsep}{4.5pt}
\scriptsize
\caption{\small Statistics on the number of tokens. $\mathcal{T}_\text{Gen}$ is the number of generated tokens during the initial answer generation, $\mathcal{T}_\text{Ver}$ is the total number of tokens verified by the streaming verifier, and $\mathcal{T}_\text{Ref}$ is the number of generated tokens during the answer refinement phase by the streaming refiner. We report the average number of tokens per answer.}
    \centering
        \begin{tabular}{l ccc ccc}
            \toprule
            \multicolumn{7}{c}{\small \textbf{ASQA}} \\
            \midrule[0.5pt]
            & \multicolumn{3}{c}{\textbf{Closed-Book}} & \multicolumn{3}{c}{\textbf{Open-Book w/ 5 Psgs}} \\
            \cmidrule(lr){2-4} \cmidrule(lr){5-7}
            Method & $\mathcal{T}_\text{Gen}$ & $\mathcal{T}_\text{Ver}$ & $\mathcal{T}_\text{Ref}$ & $\mathcal{T}_\text{Gen}$ & $\mathcal{T}_\text{Ver}$ & $\mathcal{T}_\text{Ref}$ \\
            
            \midrule[0.5pt]
            Mistral 7B & $143.8$ & $-$ & $-$ & $116.1$ & $-$ & $-$ \\
            + Full-VR & $-$ & $143.8$ & $113.6$ & $-$ & $116.1$ & $101.8$ \\
            + Streaming-VR & $-$ & $143.8$ & $68.6$ & $-$ & $116.1$ & $74.4$ \\
            
            \midrule[0.5pt]
            LLaMA-3.1 8B & $101.2$ & $-$ & $-$ & $66.9$ & $-$ & $-$ \\
            + Full-VR & $-$ & $101.2$ & $117.4$ & $-$ & $66.9$ & $106.8$ \\
            + Streaming-VR & $-$ & $101.2$ & $63.6$ & $-$ & $66.9$ & $61.9$ \\

            \midrule[0.5pt]
            GPT-4o & $100.4$ & $-$ & $-$ & $60.2$ & $-$ & $-$ \\
            + Full-VR & $-$ & $100.4$ & $107.6$ & $-$ & $60.2$ & $116.1$ \\
            + Streaming-VR & $-$ & $100.4$ & $61.9$ & $-$ & $60.2$ & $62.7$ \\

            \midrule[0.8pt]

            \multicolumn{7}{c}{\small \textbf{QuoteSum}} \\
            \midrule[0.5pt]
            & \multicolumn{3}{c}{\textbf{Zero-Shot}} & \multicolumn{3}{c}{\textbf{Five-Shots}} \\
            \cmidrule(lr){2-4} \cmidrule(lr){5-7}
            Method & $\mathcal{T}_\text{Gen}$ & $\mathcal{T}_\text{Ver}$ & $\mathcal{T}_\text{Ref}$ & $\mathcal{T}_\text{Gen}$ & $\mathcal{T}_\text{Ver}$ & $\mathcal{T}_\text{Ref}$ \\

            \midrule[0.5pt]
            Mistral 7B & $120.4$ & $-$ & $-$ & $92.5$ & $-$ & $-$ \\
            + Full-VR & $-$ & $120.4$ & $101.3$ & $-$ & $92.5$ & $72.5$ \\
            + Streaming-VR & $-$ & $120.4$ & $75.2$ & $-$ & $92.5$ & $54.9$ \\
            
            \midrule[0.5pt]
            LLaMA-3.1 8B & $161.3$ & $-$ & $-$ & $83.6$ & $-$ & $-$ \\
            + Full-VR & $-$ & $161.3$ & $154.3$ & $-$ & $83.6$ & $84.1$ \\
            + Streaming-VR & $-$ & $161.3$ & $106.1$ & $-$ & $83.6$ & $58.9$ \\

            \midrule[0.5pt]
            GPT-4o & $58.5$ & $-$ & $-$ & $65.2$ & $-$ & $-$ \\
            + Full-VR & $-$ & $58.5$ & $60.7$ & $-$ & $65.2$ & $78.9$ \\
            + Streaming-VR & $-$ & $58.5$ & $44.5$ & $-$ & $65.2$ & $45.6$ \\
            \bottomrule
        \end{tabular}
\label{tab:cost}
\end{table}

\paragraph{Streaming-VR enhances token efficiency}
In terms of efficiency, Streaming-VR offers substantial advantages over Full-VR across all models and both closed-book and open-book settings. While Full-VR refines the entire response, generating more tokens for error correction with unnecessary token refinement, Streaming-VR operates at the sentence level, refining only those sentences identified as inaccurate, resulting in significantly fewer tokens being produced. The key to Streaming-VR's efficiency lies in its ability to minimize error propagation during the generation process. By addressing inaccuracies early at the sentence level, it reduces the need for extensive revisions in subsequent stages with inefficiencies. This streamlined process leads to token savings of 39.8\% for ASQA and 31.5\% for QuoteSum on average.

\paragraph{Streaming-VR enhances time efficiency too}
We further provide a comprehensive analysis of the overall inference costs extending our evaluation beyond token efficiency in~\Cref{tab:cost}. Streaming-VR consistently produces a significantly smaller number of tokens than Full-VR, as it skips unnecessary modifications to sentences that are already correct. However, beyond the token counts, for practical deployment in real-world applications, latency plays a critical role in assessing efficiency. Specifically, latency is directly influenced by various factors other than token counts, such as the number of model invocations, execution time per call, and whether the models operate in parallel.
Compared to the na\"ive method (purely sequential generation without any verification or refinement), sentence-level correction introduces some inherent delay, as each sentence is verified and refined before proceeding to the next sentence. However, this delay is mitigated by Streaming-VR's streamlined correction mechanism, where the incorrect sentence is processed in parallel with sentence generation and verification. As a result, per-sentence verification and refinement do not accumulate linearly, keeping overall latency manageable. Importantly, the external verifier and refiner in Streaming-VR are invoked only once and operate in parallel.

We measured the latency of each method by timing the answer refinement process when an answer is determined to be incorrect. Specifically, using Mistral 7B as the verifier and LLaMA-3.1 8B as the refiner, Streaming-VR requires only an average of 3.07s, whereas Full-VR takes an average of 5.98s to refine the entire answer per question. This makes Streaming-VR approximately 1.95$\times$ faster than Full-VR while preserving answer factuality. This analysis underscores the novelty of Streaming-VR across the entire pipeline. For a more detailed explanation along with mathematical descriptions on general cases, please refer to~\Aref{app:latency}.

\begin{table}[t]
\centering
\setlength{\tabcolsep}{3pt}
\scriptsize
\caption{\small Result of Streaming-VR with \textbf{LLaMA-3.1 8B} as the response model. Models are indicated as Streaming-\{\texttt{Verifier}\}\{\texttt{Refiner}\}, where M, L and G stand for Mistral-7B, LLaMA-3.1 8B and GPT-4o, respectively.}
    \centering
        \begin{tabular}{l ccc ccc}
            \toprule
            \multicolumn{7}{c}{\small \textbf{ASQA}} \\
            \midrule[0.5pt]
            & \multicolumn{3}{c}{\textbf{Closed-Book}} & \multicolumn{3}{c}{\textbf{Open-Book w/ 5 Psgs}} \\
            \cmidrule(lr){2-4} \cmidrule(lr){5-7}
            Method & R-L & Dis-F1 & DR & R-L & Dis-F1 & DR \\
        
            \midrule[0.5pt]
            LLaMA-3.1 8B & $34.0$ & $23.7$ & $28.4$ & $36.6$ & $31.7$ & $34.1$ \\
            + Streaming-MM & $34.5$ & $23.7$ & $28.6$ & $36.2$ & $30.5$ & $33.2$ \\
            + Streaming-ML & $34.2$ & $24.3$ & $28.8$ & $36.8$ & $31.1$ & $33.8$ \\
            + Streaming-MG & $35.3$ & $29.4$ & $32.2$ & $36.8$ & $34.0$ & $35.4$ \\
            + Streaming-LG & $35.2$ & $28.3$ & $31.6$ & $36.8$ & $33.8$ & $35.3$\\
            \midrule[0.5pt]
            + \textit{Self-VR} & $34.2$ & $23.3$ & $28.2$ & $36.6$ & $31.1$ & $33.7$ \\

            \midrule[0.8pt]
            \multicolumn{7}{c}{\small \textbf{QuoteSum}} \\
            \midrule[0.5pt]
            & \multicolumn{3}{c}{\textbf{Zero-Shot}} & \multicolumn{3}{c}{\textbf{Five-Shots}} \\
            \cmidrule(lr){2-4} \cmidrule(lr){5-7}
            Method & R-L & Sem-F1 & SEMQA & R-L & Sem-F1 & SEMQA \\
        
            \midrule[0.5pt]
            LLaMA-3.1 8B & $43.3$ & $38.9$ & $41.0$ & $59.1$ & $61.2$ & $60.1$ \\
            + Streaming-MM & $39.6$ & $38.9$ & $39.3$ & $58.0$ & $61.2$ & $59.6$ \\
            + Streaming-ML & $45.2$ & $39.0$ & $41.9$ & $59.9$ & $61.7$ & $60.8$ \\
            + Streaming-MG & $47.5$ & $39.0$ & $43.0$ & $60.7$ & $62.3$ & $61.5$ \\
            + Streaming-LG & $47.7$ & $39.0$ & $43.1$ & $61.0$ & $62.3$ & $61.6$ \\
            \midrule[0.5pt]
            + \textit{Self-VR} & $42.4$ & $38.9$ & $40.6$ & $57.4$ & $61.2$ & $59.3$ \\

            \bottomrule
            \end{tabular}
\label{tab:verifier_refiner}
\end{table}

\paragraph{Verification models don't need to be bigger}
The results in \Cref{tab:verifier_refiner} show that verifier models can be effective without being large. On both tasks, Streaming-MG performs comparably to Streaming-LG, demonstrating that smaller models can still deliver significant performance gains. These findings highlight that the choice of verifiers is very robust in Streaming-VR, leading to the choice of smaller models that are resource-efficient and effective, making them valuable for real-world applications with limited computational resources.

\paragraph{Refinement models need to be bigger}
The results in~\Cref{tab:verifier_refiner} highlight the critical role of a larger and more advanced model for refinement after verification, even when the verifier is relatively small. Using Mistral 7B as verifier and refiner (Streaming-MM) results in no improvement or even degraded performance across datasets and settings.

In contrast, larger refiners yield significant gains. With LLaMA-3.1 8B as the refiner (Streaming-ML), there is a modest Dis-F1 improvement for the closed-book setting on ASQA, though handling multiple passages remains challenging. On QuoteSum, Streaming-ML achieves notable improvements in both zero- and five-shot settings, while Streaming-MM reduces answer quality. The most substantial boost comes from GPT-4o as the refiner (Streaming-MG), whose advanced reasoning capabilities drive superior performance in both RAG and ICL settings. These results confirm the importance of using a refiner larger than the response model for coherent and accurate answers.

\paragraph{LLMs still struggle with intrinsic self-correction}
\label{self-correction}
Additionally, we conduct some experiments to evaluate the efficacy of self-verification and self-refinement within the Streaming-VR pipeline, utilizing only LLaMA-3.1 8B for backbone, verifier and refiner models. In~\Cref{tab:verifier_refiner}, the rows of \textit{Self-VR} (Self-Verification and Refinement; \textit{i.e.,} Streaming-LL) illustrate that LLMs continue to face challenges with intrinsic self-correction with some performance drops. This result strengthens the conclusions drawn by \citet{self-correct-reasoning},  which also have demonstrated that intrinsic self-correction, an approach that model attempts to rectify its initial responses using only its inherent capabilities without external feedback, degrades the response quality.

\begin{figure}[t]
    \Large
    \centering
    \resizebox{0.975\linewidth}{!}{
    \begin{tabular}{cc}
        \includegraphics[width=1.1\linewidth]{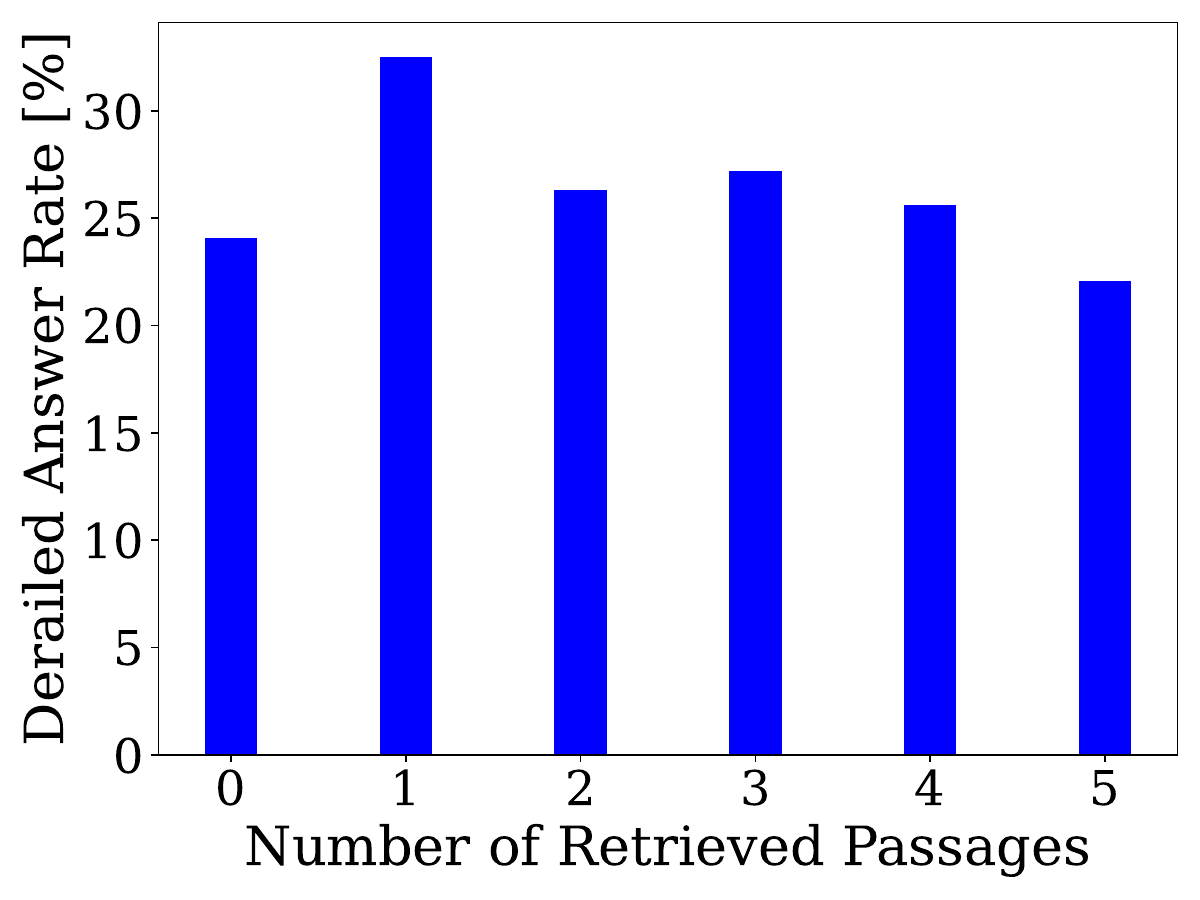} &
        \includegraphics[width=1.1\linewidth]{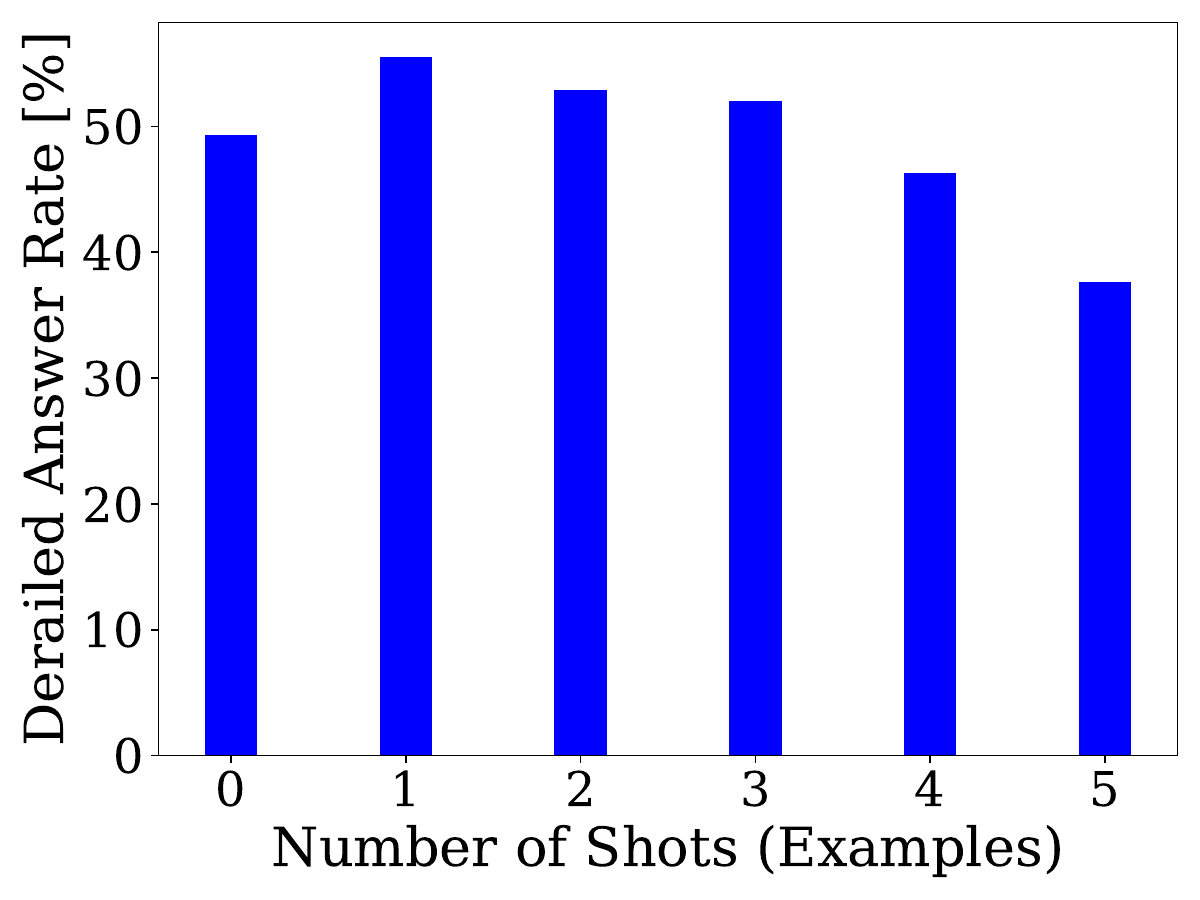} \\
        \textbf{\phantom{,,,,,,,}(a) ASQA} &
        {\textbf{\phantom{,,,,,,,}(b) QuoteSum}} \\
    \end{tabular}
    }
\caption{\small The ratio of derailed answers to incorrect answers. \textbf{(a)}: The rate of derailed answers on \textbf{ASQA}. \textbf{(b)}: The rate of derailed answers on \textbf{QuoteSum}.}
\label{fig:derailed_rate}
\end{figure}

\paragraph{Errors in the middle derail the entire answer}
As \citet{snowball} point out that the mistakes or hallucinations in the middle of the answer can skew the whole response, we report the statistics of model-generated answers with the rate of derailed answers on each dataset. Specifically, the answers are generated by LLaMA-3.1 8B and verified by the finetuned streaming verifier as before. The rate of derailed answers is the ratio of the number of `answers composed of false sentences in sequence from the first erroneous sentence to the last one' to `false answers if at least one of their sentences is identified as false'. Results in \Cref{fig:derailed_rate} for ASQA and QuoteSum are 26.3\% and 48.9\% on average across different settings for RAG and ICL, respectively. Therefore, they highlight the importance of Streaming-VR to prevent derailed responses.

\section{Conclusion}
In this paper, we introduced Streaming-VR, a novel approach aimed at improving the accuracy and efficiency in language model text generation. Unlike traditional methods solely relying on the final response, Streaming-VR performs real-time verification and correction of erroneous token sequences as they are being produced, with external models simultaneously with answer generation. This prevents error propagation in the early stage and reduces the errors at the end by minimizing the likelihood of compounding inaccuracies, then significantly enhances the efficiency of answer refinement. Extensive experiments for two different question-answering datasets have clearly demonstrated that Streaming-VR consistently achieves remarkably higher efficiency without compromising accuracy.

\section*{Limitations}
Despite the improvements from Streaming-VR, which enhances both the efficiency and effectiveness of verification and refinement in language model text generation by intervening during intermediate answer generation, there remain promising opportunities for enhancing the answer verifier. Specifically, the primary challenge is the lack of dedicated datasets for answer verification, particularly those suited for real-time scenarios. To address this, we automatically augmented data by paraphrasing sentences or introducing errors by an LLM. However, while effective, this approach carries the risk of mislabeling. Therefore, future work could focus on developing new datasets that are carefully annotated with a diverse range of answers ensuring more accurate verification and reducing the risk of incorrect labeling. Additionally, we can further extend these datasets to include fine-grained labels for multiple classes, rather than just binary ones, to accommodate different types of errors and apply adaptive strategies for subsequent refinement after verification.

\section*{Ethics Statement}
In our research, we use publicly available question-answering (QA) datasets to evaluate the effectiveness and applicability of Streaming-VR in real-world scenarios. The language model we employ may inadvertently reflect biases embedded in its training data, resulting in outputs that perpetuate racism, sexism, or other forms of discrimination. Such biases can manifest even in contexts that appear neutral, highlighting the need for proactive bias detection and mitigation strategies. Moreover, harmful inputs might lead to the retrieval of offensive information or the generation of inappropriate responses by the language models. This presents a significant risk that we must recognize and address. To mitigate these issues, it is crucial to develop methods for detecting and managing offensive, inappropriate, or biased content in both user inputs and the documents retrieved within our retrieval-augmented framework. We view this as a critical area for future research because minimizing the risk of biased or harmful outputs is essential for the safe and ethical deployment of QA systems.

\section*{Acknolwedgments}
This work was supported by the Institute for Information \& Communications Technology Planning \& Evaluation (IITP) grant funded by the Korea government (MSIT) (RS-2019-II190075, Artificial Intelligence Graduate School Program (KAIST), and RS-2022-II220713, Meta-learning Applicable to Real-world Problems), the National Research Foundation of Korea (NRF) grant funded by the Korea government (MSIT) (RS-2023-00256259), the grant of the Korea Machine Learning Ledger Orchestration for Drug Discovery Project (K-MELLODDY) funded by the Ministry of Health \& Welfare and the Ministry of Science and ICT, Republic of Korea (RS2024-00460870), Institute of Information \& Communications Technology Planning \& Evaluation (IITP) with the grant funded by the Ministry of Science and ICT (MSIT) of the Republic of Korea in connection with the Global AI Frontier Lab International Collaborative Research (RS-2024-00469482 \& RS-2024-00509279), the Artificial Intelligence Industrial Convergence Cluster Development Project funded by the Ministry of Science and ICT (MSIT, Korea) \& Gwangju Metropolitan City, and i-Scream Media.

\bibliography{custom}

\clearpage
\appendix

\section{Implementation Details}
\label{implementation details}

\paragraph{Models}
In our experiments, we employ two open-source LLMs \href{https://huggingface.co/mistralai/Mistral-7B-Instruct-v0.3}{Mistral 7B}~\citep{mistral} and \href{https://huggingface.co/meta-llama/Llama-3.1-8B}{LLaMA-3.1 8B}~\citep{llama3} via \href{https://huggingface.co/}{Hugging Face Hub}~\citep{huggingface} and GPT-4o~\citep{gpt4} which is accessible via \href{https://platform.openai.com/docs/models/gpt-4o}{OpenAI} API, representing relatively small, medium, and large models, respectively.
Here, these models are never fine-tuned or further trained except for their roles in verification. For the overall Streaming-VR pipeline, LLaMA-3.1 8B functions as the primary backbone language model to generate answers for given queries, while all three models are employed for verification or refinement for experiments.

\paragraph{Streaming Verifier}
\label{streaming verifier}

We fine-tuned Mistral 7B and LLaMA-3.1 8B as verifiers using augmented training data from ASQA and QuoteSum datasets. Each verifier was trained for five epochs on its respective training set, with a learning rate of 1e-5 and the AdamW~\citep{adamw} optimizer. To generate augmented data for false-labeled sentences, we embedded fake information into true sentences using GPT-4o, adjusting the temperature to $0.3$, $0.5$, and $0.7$ to create diverse forms of inaccuracies. Rather than synthesizing entirely new sentences with large language models, which risk introducing unrelated hallucinations, we adopted this targeted augmentation strategy as a more reliable approach. This method proved highly effective in training verifiers to identify hallucinations, delivering exceptional results that highlight the importance of Streaming-VR in improving efficiency while preserving answer quality. The specific prompt used to generate incorrect information for each sentence (\texttt{Sentence}) in the provided answer (\texttt{Answer}) to a given question (\texttt{Question}) is detailed in the prompt box as follows.

\begin{Box1}{\small Prompt}\label{app:augmentation_prompt}
\small
You will be given a question (Q) with its corresponding answer paragraph (A) that may be incomplete and a sentence (S) following the paragraph.\\

Q: \texttt{\{Question\}}\\
A: \texttt{\{Answer\}}\\
S: \texttt{\{Sentence\}}\\

You should modify the given sentence S, into a plausible lie by inserting some wrong information. Just return only the modified sentence (S) itself.
\end{Box1}

\begin{table}[h!]
\centering
\setlength{\tabcolsep}{7pt}
\small
\caption{\small The results of Streaming Verifier finetuned on the train set for each dataset. We report the test accuracy of the verifiers along with a random classifier.}
    \centering
        \begin{tabular}{l cc}
            \toprule
            \textbf{Method} & \textbf{ASQA} & \textbf{QuoteSum} \\
            \midrule[0.5pt]
            Random & $49.6$ & $50.3$ \\
            Mistral 7B & $86.8$ & $81.7$ \\
            LLaMA-3.1 8B & $86.7$ & $93.0$ \\
            \bottomrule
        \end{tabular}
\label{tab:verifier_acc}
\end{table}

The final test results of the finetuned verifiers used in the experiments, including a Random baseline that selects verification results arbitrarily, are presented in \Cref{tab:verifier_acc}. For the entire pipeline, we establish a constraint that prohibits the use of any other verifier models larger than the answer generation model. This decision is based on the principle that the verifier should not exceed the capabilities of the response model, as the verifier serves merely as a supplementary tool for identifying mistakes. This reflects our considerations regarding computational costs and efficiency.

\section{Additional Experimental Results}

\begin{table}[h]
\centering
\setlength{\tabcolsep}{5pt}
\scriptsize
\caption{\small Result of the Self-RAG baseline on ASQA.}
    \centering
    \begin{tabular}{l ccc ccc}
        \toprule
        \multicolumn{7}{c}{\small \textbf{ASQA}} \\
        \midrule[0.5pt]
        & \multicolumn{3}{c}{\textbf{Closed-Book}} & \multicolumn{3}{c}{\textbf{Open-Book w/ 5 Psgs}} \\
        \cmidrule(lr){2-4} \cmidrule(lr){5-7}
        Method & R-L & Dis-F1 & DR & R-L & Dis-F1 & DR \\
        
        \midrule[0.7pt]
        Self-RAG 7B & $22.5$ & $13.0$ & $17.1$ & $32.6$ & $27.5$ & $29.9$ \\
        Self-RAG 13B & $21.0$ & $15.1$ & $17.9$ & $34.1$ & $29.7$ & $31.8$ \\

        \midrule[0.5pt]
        Mistral 7B & $33.6$ & $20.7$ & $26.4$ & $36.4$ & $31.2$ & $33.7$ \\
        + Streaming-VR & $35.2$ & $29.6$ & $32.3$ & $36.9$ & $33.7$ & $35.3$ \\
        
        \bottomrule
    \end{tabular}
\label{tab:self-rag}
\end{table}

\paragraph{Additional Baselines}
In \Cref{tab:self-rag}, we present additional results for other baselines on ASQA, Self-RAG~\citep{self-rag}, one of the most representative methods with their trained models publicly available. Self-RAG performs on-demand retrieval of external information via a specialized retrieval token, followed by a critique of the generated output to refine it. When we compare the baseline results with the similar model size, Mistral 7B demonstrates significantly superior performance to Self-RAG even without the help of refinement.

\section{Time-Efficiency of Streaming-VR}
\label{app:latency}

\paragraph{Streaming-VR is always faster than Full-VR}
The latencies of Full-VR ($t_\text{F}$) and Streaming-VR ($t_\text{S}$) can be calculated as follow:
\begin{align}
    t_{\text{F}} &= t_{\text{Ver}} + \mathcal{T}_{\text{Ref}}^{\text{F}} \times t_{\text{Ref}} \notag \\
    t_{\text{S}} &= N \times t_{\text{Ver}} + \mathcal{T}_{\text{Ref}}^{\text{S}} \times t_{\text{Ref}} \notag
\end{align}
Here, we assume that the inference time of the verification model per call, $t_{\text{Ver}}$, is the same regardless of the input length for simple comparison, and $t_{\text{Ref}}$ denotes the token generation time per token of the refiner. Since the verifier does not generate tokens, it follows that $t_{\text{Ver}} < t_{\text{Ref}}$. Furthermore, as shown in~\Cref{tab:cost}, with an average of seven sentences per answer $(N=4)$, we observe that $N + \mathcal{T}_{\text{Ref}}^{\text{S}} < \mathcal{T}_{\text{Ref}}^{\text{F}}$. Consequently, we can conclude that $t_{\text{S}} < t_{\text{F}}$ due to the following inequality:
\begin{align}
    t_{\text{S}} &< \left(N + \mathcal{T}_{\text{Ref}}^{\text{S}} \right) \times t_{\text{Ref}}
    < \mathcal{T}_{\text{Ref}}^{\text{F}} \times t_{\text{Ref}} < t_{\text{F}} \notag
\end{align}
When Streaming-VR is applied in real-world scenarios, where the verifier and refiner operate simultaneously alongside the answering model, the latency of Streaming-VR is updated to $t_{\text{S}}^{\text{real}}$ as:
\begin{align}
    t_{\text{S}}^{\text{real}} &= \max \left\{t_{\text{Ver}}, \frac{\mathcal{T}_{\text{Ref}}^{\text{S}}}{N} \times t_{\text{Ref}} \right\} \notag
\end{align}
because it only depends on the time for the last sentence to be verified (and may be refined).
So it demonstrates that Streaming-VR achieves significantly lower latency compared to Full-VR, as $t_{\text{S}}^{\text{real}} \ll t_{\text{S}} < t_{\text{F}}$. As a result, comparing the number of tokens generated during refinement is sufficient to analyze the overall latency of both methods.

\section{GPU memory}
In a practical deployment setting, Full-VR and Streaming-VR exhibit the same maximum GPU memory usage. To be specific, as the maximum number of tokens that Full-VR and Streaming-VR should handle is exactly the same, which depends on the generated (and refined) answer from the original response model. Quantitatively, when using LLaMA-3.1 8B for answer generation and Mistral 7B for verification, it requires approximately 43,900 (16,035 + 27,865) MB of GPU memory.

\section{Case Study}
\begin{table}[h]
\centering
\setlength{\tabcolsep}{3pt}
\small
\caption{\small Case Study using LLaMA-3.1 8B as the original answer generation model, Mistral 7B as the verifier, and GPT-4o as the refiner.}
    \centering
    \resizebox{0.47\textwidth}{!}{
        \begin{tabular}{p{0.14\textwidth} p{0.4\textwidth}}
            \toprule
            \multicolumn{2}{c}{\textbf{\normalsize Q: Who does sam marry in foyle's war?}}\\
            \midrule[0.7pt]
            \normalsize \textbf{Method} & \normalsize \textbf{Answer}\\
            \midrule[0.7pt]
            LLaMA-3.1 8B & In the \textbf{\textcolor{red}{BBC}} drama series Foyle's War, there are actually two main characters who get married. The first one is Sam Stewart, played by Michael Kitchen, who is the main character, David Foyle. However, Sam Stewart is a love interest and also a major character in the series. Sam Stewart marries a \textbf{\textcolor{red}{woman named Kate Baxter}}, but then also falls in love with a \textbf{\textcolor{red}{woman named Samantha Stewart}} who is a secretary at the police station and they get married.\\
            
            \midrule[0.5pt]
            + Full-VR & In the \textbf{\textcolor{red}{BBC}} drama series Foyle's War, Sam Stewart, played by Honeysuckle Weeks, is a major character who works as a driver for Detective Chief Superintendent Christopher Foyle. Sam Stewart marries \textbf{\textcolor{blue}{Adam Wainwright}}, a former Bletchley Park codebreaker who becomes a local politician. Their relationship evolves as they navigate post-war challenges, with Sam balancing her duties as a wife and her continuing involvement in Foyle's work.\\
            
            \midrule[0.5pt]
            + \textbf{Streaming-VR} & In the \textbf{\textcolor{blue}{ITV}} drama series Foyle's War, Sam marries Adam Wainwright, a former Bletchley Park codebreaker. Sam Stewart, played by Honeysuckle Weeks, is the driver for the main character, Christopher Foyle. However, Sam Stewart is a former driver for Christopher Foyle and also a major character. Sam Stewart marries a \textbf{\textcolor{blue}{man named Adam Wainwright}}, a former Bletchley Park codebreaker, and they start a life together after the war ends, with Sam continuing her role as a prominent character.\\
            
            \bottomrule
        \end{tabular}
    }
\label{tab:case_study}
\end{table}

We have conducted case studies to examine the advantages of Streaming-VR in greater detail. In~\Cref{tab:case_study}, we present the results of various methods for the query: \textit{"Who does sam marry in foyle's war?"}, a British detective drama television series broadcast on ITV. The answer to the question is \textit{Adam Wainwright}. However, this question introduces ambiguity because the answer could refer either to \textit{Max Brown}, who portrayed \textit{Adam Wainwright} in Season 6, or \textit{Daniel Weyman}, who was recast as \textit{Adam Wainwright} in Seasons 7 and 8.

As shown in~\Cref{tab:case_study}, the original answer generated by the na\"ive LLaMA-3.1 8B model reflects the model's difficulty in understanding due to grammatical errors and ambiguous information. However, when using the methods that enable the refinement through the verification (with \textcolor{red}{red} to be \texttt{False} and \textcolor{blue}{blue} to be \texttt{True}), Full-VR still contains an error regarding the name of the broadcasting center, whereas Streaming-VR successfully corrects every error in the answer. Also, we observe that the answer generated and refined by Streaming-VR preserves the structure of the original answer, whereas Full-VR generates an entirely new response, significantly altering the sentence structure.

\end{document}